\documentclass[letterpaper]{article} 
\usepackage[draft]{aaai2026}  
\usepackage{times}  
\usepackage{helvet}  
\usepackage{courier}  
\usepackage[hyphens]{url}  
\usepackage{graphicx} 
\usepackage{natbib}  
\usepackage{caption} 
\usepackage{graphicx}
\usepackage{amsmath}
\usepackage{physics}
\usepackage{amssymb}
\usepackage{booktabs}
\usepackage[utf8]{inputenc} 
\usepackage[T1]{fontenc}    
\usepackage{url}            
\usepackage{booktabs}       
\usepackage{amsfonts}       
\usepackage{nicefrac}       
\usepackage{microtype}      
\usepackage{xcolor}         
\usepackage{array}
\usepackage{amssymb}
\usepackage{latexsym}
\usepackage{amsmath,amssymb}
\usepackage{graphicx}
\usepackage{multirow}
\usepackage{adjustbox}
\usepackage{makecell}

\newlength{\commentWidth}
\newcommand{\vect}[1]{\mathbf{#1}}

\title{LaMoGen: Laban Movement-Guided Diffusion for Text-to-Motion Generation}
\author{
	Heechang Kim\textsuperscript{\rm 1} \qquad \stepcounter{footnote}
	Gwanghyun Kim\textsuperscript{\rm 1} \qquad \stepcounter{footnote}
	Se Young Chun\textsuperscript{\rm 1,2}\thanks{Corresponding author.} \\
}
\affiliations{
    \textsuperscript{\rm 1}Dept. of Electrical and Computer Engineering, \textsuperscript{\rm 2}INMC \& IPAI\\
    Seoul National University, Republic of Korea\\
    {\tt\small \{heechang.kim, gwang.kim, sychun\}@snu.ac.kr}
}

\begin{document}

\maketitle

\begin{abstract}
Diverse human motion generation is an increasingly important task, having various applications in computer vision, human-computer interaction and animation. While text-to-motion synthesis using diffusion models has shown success in generating high-quality motions, achieving fine-grained expressive motion control remains a significant challenge. This is due to the lack of motion style diversity in datasets and the difficulty of expressing quantitative characteristics in natural language. Laban movement analysis has been widely used by dance experts to express the details of motion including motion quality as consistent as possible. Inspired by that, this work aims for interpretable and expressive control of human motion generation by seamlessly integrating the quantification methods of Laban Effort and Shape components into the text-guided motion generation models. Our proposed zero-shot, inference-time optimization method guides the motion generation model to have desired Laban Effort and Shape components without any additional motion data by updating the text embedding of pretrained diffusion models during the sampling step. We demonstrate that our approach yields diverse expressive motion qualities while preserving motion identity by successfully manipulating motion attributes according to target Laban tags.
\end{abstract}

\section{Introduction}
\label{sec1_introduction}
The generation of realistic and controllable human motion~\cite{holden2016deep} is a pivotal task with wide-ranging applications in human-computer interaction~\cite{kim2009stable}, virtual reality~\cite{mousas2018performance}, and computer animation~\cite{gleicher2001motion}. The field has recently seen a paradigm shift towards text-to-motion synthesis~\cite{petrovich2022temos}, leveraging the expressive power of natural language~\cite{ahuja2019language2pose}. Early attempts~\cite{guo2020action2motion} were often limited to generating motions from a small set of predefined action classes. In contrast, modern state-of-the-art methods~\cite{tevet2022human, zhang2022motiondiffuse}, predominantly based on diffusion models trained on large-scale datasets like HumanML3D~\cite{guo2022generating}, can generate diverse and high-quality motions directly from complex textual descriptions.

\begin{figure}[!t]
	\centering
	\includegraphics[width=1.0\linewidth]{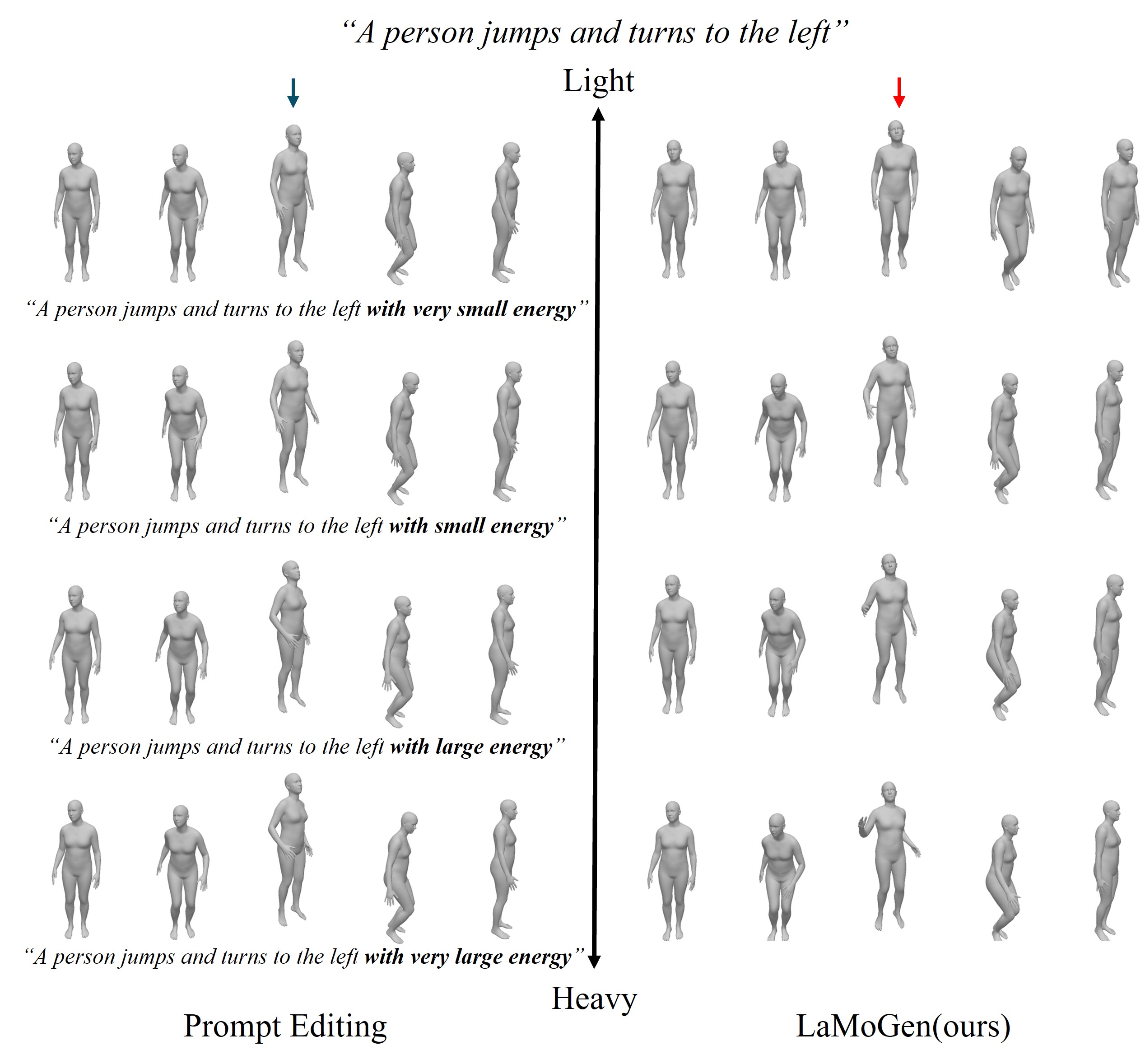}
	\vspace{-0.8em}
	\caption{Existing text to motion generation methods ~\cite{zhang2022motiondiffuse} cannot control the extents of motion by prompt, while our method can control them. In the third frame marked with arrows, existing methods (blue arrow) generates same motion, while our method (red arrow) generate diverse motion with increasing energy.}
	\label{fig1}
\end{figure}

Despite this progress, a significant challenge remains: achieving fine-grained control over the stylistic and expressive qualities of the generated motion. This limitation stems from two fundamental issues. First, even the largest public motion datasets~\cite{lin2023motionx} are orders of magnitude smaller than the text-image corpora~\cite{imagenet} used for image generation. This data scarcity limits the model's capability to capture subtle expressive nuances, often causing it to fail in rendering specific details including motion quality. Second, natural language, while powerful for semantic description, is often ambiguous and insufficient for specifying the quantitative degree of non-verbal attributes like energy, fluidity, or tension. This creates a fundamental barrier to generating a motion that precisely matches a user's intention.

To address this gap, we turn to the domain of performance arts and leverage Laban Movement Analysis (LMA)~\cite{laban1971mastery} as a structured, quantitative framework for motion description. LMA provides a formal vocabulary for non-verbal motion, particularly through its Effort and Shape components. The Laban Effort category describes the subtle, dynamic qualities of movement through four factors (Space, Weight, Time, and Flow), while the Shape category describes how the body changes its form in space. Prior works have successfully quantified these components for full-body motion analysis and affective generation~\cite{samadani2013laban, samadani2020affective}, establishing a solid mathematical basis for their use.

Motivated by these advancements on LMA and its mathematical basis, we propose \textbf{La}ban movement-guided diffusion for text to \textbf{Mo}tion \textbf{Gen}eration (\textbf{LaMoGen}), a novel framework for fine-grained controllable motion generation. Our method augments the standard text prompt with quantitative Laban Effort and Shape tags to enable detailed control over expressive motion characteristics. The core of our approach is a zero-shot, inference-time optimization strategy that guides a pretrained diffusion model, such as MotionDiffuse~\cite{zhang2022motiondiffuse}, to produce motions that adhere to both the semantic text description and the expressive LMA targets.

To achieve this, we first develop a set of differentiable functions that quantify LMA features from motion data, adapting the equations from prior work~\cite{samadani2020affective} to be suitable for gradient-based updates. A key challenge is that training a model with LMA supervision from scratch is infeasible due to the lack of large-scale motion datasets with ground-truth Laban annotations. Our method circumvents this by operating exclusively at inference time. We introduce the ``Laban loss'' that measures the discrepancy between the LMA features of the intermediate generated motion and the desired target values. During the reverse diffusion process, the gradient of this loss with respect to the conditional text embedding is computed and then the embedding is updated to minimize the discrepancy. This guidance step, applied throughout sampling, steers the generation towards the intended expressive motion quality while preserving the motion's core identity.

Our quantitative evaluation demonstrates that integrating Laban Effort and Shape tags into text-to-motion generations improves the controllability over expressive motion quality, outperforming baseline methods such as explicit textual descriptions or post-generation manipulation, particularly in terms of diagonality metrics. Furthermore, our method maintains competitive performance on standard benchmarks in FID and R-precision.
The contributions of this work are summarized as follows:
\begin{enumerate}
    \item We introduce the first approach to leverage quantitative Laban Movement Analysis for guiding pretrained text-to-motion diffusion models, enabling fine-grained and interpretable expressive control.
    
    \item We propose a zero-shot, inference-time optimization method that requires no additional training or annotated data, making it a readily applicable, plug-and-play module for existing models.
    
    \item We demonstrate through quantitative metrics that our method successfully manipulates motion attributes according to target Laban tags with minimal degradation in the original motion's identity. 
\end{enumerate}

\section{Related Work}
\subsection{Controllable Generative Models}
Diffusion models~\cite{ho2020denoising} have emerged as the leading paradigm for high-quality content generation. Their effectiveness is significantly enhanced by guidance techniques that steer the generation process. Classifier-Free Guidance~\cite{ho2022classifier}, in particular, is widely used to enable conditioning on inputs like text by extrapolating between a conditional and an unconditional model prediction at inference time.

Building on this foundation, recent works focus on enabling more fine-grained, zero-shot control without costly model retraining. These methods typically operate by manipulating the model's internal representations, such as cross-attention maps, during the denoising process. For example, ControlNet~\cite{zhang2023adding} allows for conditioning on various spatial inputs like edge maps or human pose skeletons, while other works enable control via bounding boxes~\cite{xie2023boxdiff}. These approaches inform our own, as we similarly apply guidance at inference time. However, where they focus on controlling spatial layout, our work targets the control of high-level expressive qualities and temporal dynamics of motion.
\subsection{Text-to-Motion Generation}
The application of diffusion models to text-to-motion synthesis has yielded remarkable progress. Foundational works like MDM~\cite{tevet2022human} and MotionDiffuse~\cite{zhang2022motiondiffuse} established that Transformer-based architectures trained on large-scale motion capture datasets, such as HumanML3D~\cite{guo2022generating}, can produce diverse and realistic motions from text prompts. Subsequent works have focused on improving diversity through retrieval augmentation~\cite{zhang2023remodiffuse} or by scaling up training data~\cite{azadi2023make}.

A significant line of research has concentrated on adding explicit control to motion generation. One category of control addresses low-level kinematic or physical constraints. For instance, PhysDiff~\cite{yuan2022physdiff} incorporates physical plausibility, while methods like Omnicontrol~\cite{xie2024omnicontrol} and TLControl~\cite{wan2023tlcontrol} allow for precise control over joint trajectories. A second category addresses high-level stylistic control. SMooDi~\cite{zhong2024smoodi}, for example, enables the transfer of an overall motion style from a reference clip to a newly generated one.

However, these two categories of control leave a critical gap. Kinematic control does not address expressive quality, while example-based style transfer requires the user to find a suitable reference motion, and the transferred style often remains a monolithic "black box." Our work introduces a third, complementary modality of control: a descriptive, component-based approach. We empower users to sculpt the motion's expressive composition using intuitive, semantic descriptors from Laban Movement Analysis, offering disentangled control without the need for a reference example.
\subsection{Laban Movement Analysis in Motion Synthesis}
Laban Movement Analysis (LMA) provides a rich, qualitative vocabulary for describing human movement, making it an attractive framework for expressive motion synthesis. Prior research has explored its use in this context. For instance, Samadani et al.\cite{samadani2020affective} used Hidden Markov Models conditioned on LMA factors to generate expressive gestures. Similarly, LaViers et al.\cite{LaViers2016} formulated Laban Efforts as a cost function within an optimal control framework to generate expressive robotic motions.

These pioneering efforts, however, were built on classical generative or control-theoretic methods (e.g., HMMs, FSMs), which are less flexible and produce lower-fidelity results compared to modern deep generative models. To the best of our knowledge, no prior work has successfully integrated the high-level, descriptive power of LMA with the generative quality and free-form conditioning capabilities of text-to-motion diffusion models. Our work is the first to bridge this gap, proposing a novel method to translate abstract Laban concepts into a differentiable guidance signal for diffusion-based motion synthesis.

\section{Method}
\begin{figure*}[!t]
	\centering
	\includegraphics[width=0.95\linewidth]{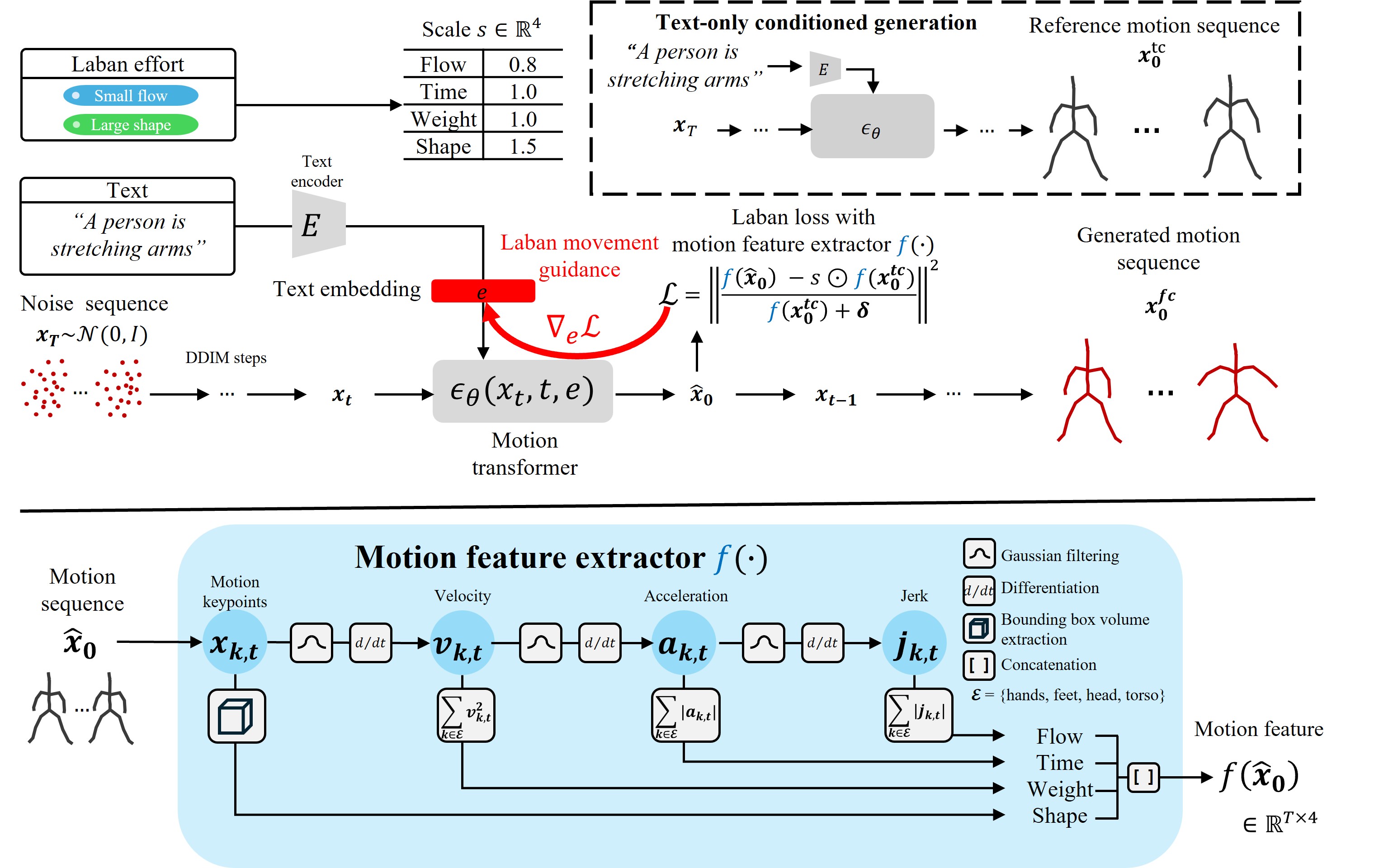}
	\vspace{-0.3em}
	\caption{The overall pipeline of our proposed method. Our framework first generates a text-conditioned baseline motion to establish content-aware target features. It then performs a second, guided generation pass, iteratively optimizing the text embedding to match the desired Laban characteristics. The differentiable motion feature extraction process is detailed at the bottom.}
	\label{fig:method_pipeline}
\end{figure*}

Our objective is to control a text-to-motion diffusion model using intuitive, Laban-based descriptors. To this end, we propose a novel framework, illustrated in Figure~\ref{fig:method_pipeline}, that consists of three key components. First, we introduce a differentiable formulation of Laban Effort and Shape features, making them suitable for gradient-based optimization. Second, we design a two-step generation pipeline that establishes a content-aware target for these features by first generating a baseline motion from the text prompt. Finally, we introduce a guidance mechanism that steers the diffusion process by iteratively optimizing the text embedding at each sampling step to align the generated motion with the target Laban characteristics.

\subsection{Preliminaries}
\subsubsection{Diffusion-Based Motion Generation}
Let $x_0$ be a motion sequence. A text-conditioned, $\epsilon$-prediction diffusion model, $\epsilon_\theta(x_t, t, c)$, is trained to predict the noise $\epsilon$ added to a sample $x_t$ at timestep $t$, given a text condition $c$. The training objective is:
\begin{align}
    \mathcal{L} = \mathbb{E}_{x_0, c, \epsilon, t}  \|\epsilon_\theta(x_t, t, c) - \epsilon\|^2
\end{align}
where $x_t = \sqrt{\bar{\alpha}_t}x_0 + \sqrt{1-\bar{\alpha}_t}\epsilon$ and $\bar{\alpha}_t$ is the noise schedule. At inference, a motion is generated by iteratively denoising a random sample $x_T \sim \mathcal{N}(0, \mathbf{I})$.

Our generation process leverages the Denoising Diffusion Implicit Model (DDIM)~\cite{song2020denoising}, a deterministic variant of the reverse process. DDIM provides faster sampling and, crucially for our two-step approach, ensures that a fixed initial noise $x_T$ yields a deterministic output. The DDIM update rule to predict the clean sample $\hat{x}_0$ and the next state $x_{t-1}$ is:
\begin{align}
	\hat{x}_0 &= \frac{1}{\sqrt{\bar{\alpha}_t}} (x_t - \sqrt{1-\bar{\alpha}_t} \cdot \epsilon_\theta(x_t, t, c)) \\
	x_{t-1} &= \sqrt{\bar{\alpha}_{t-1}} \hat{x}_0 + \sqrt{1 - \bar{\alpha}_{t-1}} \cdot \epsilon_\theta(x_t, t, c)
\end{align}
The final generated motion $x_0$ is a deterministic function of the initial latent variable $z = x_T$.

\subsubsection{Laban Effort and Shape Quantification}
\label{sec:laban_quantification}
We build upon the Laban Effort and Shape quantifications proposed by Samadani et al.~\cite{samadani2020affective}. We focus on four perceptually significant components from their work: Weight Effort, Time Effort, Flow Effort, and Shape Flow. In their original formulation, these are defined as the maximum value of a physical quantity over a motion sequence. For a sequence with end-effectors $\mathcal{E} = \{\text{hands}, \text{feet}, \text{head}, \text{torso}\}$, the components are defined using kinematic quantities $\vect{v}_{k,t}, \vect{a}_{k,t}, \vect{j}_{k,t}$ which are the velocity, acceleration, and jerk for a joint $k$ in time $t$.

\begin{itemize}
    \item \textbf{Weight} (from kinetic energy): $\max_t \sum_{k \in \mathcal{E}} \|\vect{v}_{k,t}\|^2$. \\
    When Weight is small (large), motion is said to be light (strong).
    \item \textbf{Time} (from acceleration): $\max_t \sum_{k \in \mathcal{E}} \|\vect{a}_{k,t}\|$. \\
    When Time is small (large), motion is said to be sustained (sudden).
    \item \textbf{Flow} (from jerk): $\max_t\sum_{k \in \mathcal{E}} \|\vect{j}_{k,t}\|$. \\
    When Flow is small (large), motion is said to be bound (free).
    \item \textbf{Shape} (from body volume): $\max_t V_t$, where $V_t$ is the volume of the skeleton's 3D bounding box at frame $t$. \\
    When Shape is small (large), motion is said to be near (far).
\end{itemize}

\subsection{Differentiable Laban Feature Extraction}
\label{sec:laban_features}
The original Laban definitions rely on the non-differentiable $\max$ operator, rendering them unsuitable for gradient-based guidance. A naive differentiable alternative, such as temporal mean, fails to capture the perceptually salient peak moments that define these qualities.

To overcome this, we reformulate the Laban features as the complete time-series of their underlying physical quantities. Let $E_t$, $A_t$, $J_t$, and $V_t$ be the instantaneous values for Weight, Time, Flow, and Shape at frame $t$. Our feature extraction function $f$ maps an input motion $x_0 \in \mathbb{R}^{T \times D}$ to a multi-channel feature sequence:
\begin{equation}
	f(x_0) = [E_t, A_t, J_t, V_t]_{t=1}^T \in \mathbb{R}^{T \times 4}
\end{equation}
where $T$ is the motion length and $D$ is the pose dimension. By constraining this entire time-series, we can proportionally scale the features at every frame. This indirectly manipulates the peak value, effectively controlling the original Laban definitions in a differentiable manner. To ensure numerical stability and mitigate high-frequency noise amplified by finite difference calculations, we apply a Gaussian smoothing filter to the input kinematics before computing derivatives.

\subsection{Two-Step Guided Generation}
\label{sec:two_step_generation}
A key challenge in expressive motion control is defining a meaningful target value for expressive features. An absolute target (e.g., "Weight = 50") is ill-posed, as the appropriate value is highly dependent on the motion's content (e.g., "jumping" vs. "tiptoeing"). Our two-step approach resolves this ambiguity by setting a relative, content-aware target.

\subsubsection{Step 1: Baseline Motion and Target Definition}
First, we generate a baseline motion $x_0^{\text{tc}}$ conditioned solely on the text prompt $c$, using the deterministic DDIM sampler. This motion serves as a content-specific reference. We extract its Laban feature time-series, $f(x_0^{\text{tc}})$. The final target feature, $f_{\text{target}}$, is defined by scaling this baseline. User-provided tags (e.g., "strong", "sudden") are mapped to a 4-dimensional scaling vector $\mathbf{s}$. The target is then computed as:
\begin{equation}
    f_{\text{target}} = \mathbf{s} \odot f(x_0^{\text{tc}})
\end{equation}
where $\odot$ denotes element-wise multiplication, broadcast across the time dimension. Users can also specify the scaling vector $\mathbf{s}$ directly for fine-grained control.

\subsubsection{Step 2: Laban-Guided Generation}
In the second step, we generate the final, fully-conditioned motion $x_0^{\text{fc}}$. To preserve the content and structure from the baseline, we start from the same initial latent noise $z = x_T$ used in Step 1.

Guidance is applied during the DDIM sampling process by minimizing a loss function that measures the relative difference between the predicted motion's Laban features and the target:
\begin{equation}\label{eq:laban_loss}
\mathcal{L}_{\text{Laban}}(\hat{x}_0) = \left\| \frac{f(\hat{x}_0) - f_{\text{target}}}{f(x_0^{\text{tc}}) + \delta} \right\|^2
\end{equation}
where $\hat{x}_0$ is the predicted clean motion at a given diffusion step and $\delta$ is a small constant for stability. This relative loss ensures that guidance is proportional to the baseline motion's characteristics.

We achieve this guidance by treating the projected text embedding, $e \in \mathbb{R}^{d_{\text{proj}}}$, as an optimizable parameter. At each sampling step $t$, we compute the gradient of the Laban loss with respect to the embedding, $\nabla_e \mathcal{L}_{\text{Laban}}$, and update $e$ using an Adam optimizer. The updated embedding $e'$ is then used for the denoising step that produces $x_{t-1}$.

\begin{figure*}[!t]
	\centering
	\includegraphics[width=0.9\linewidth]{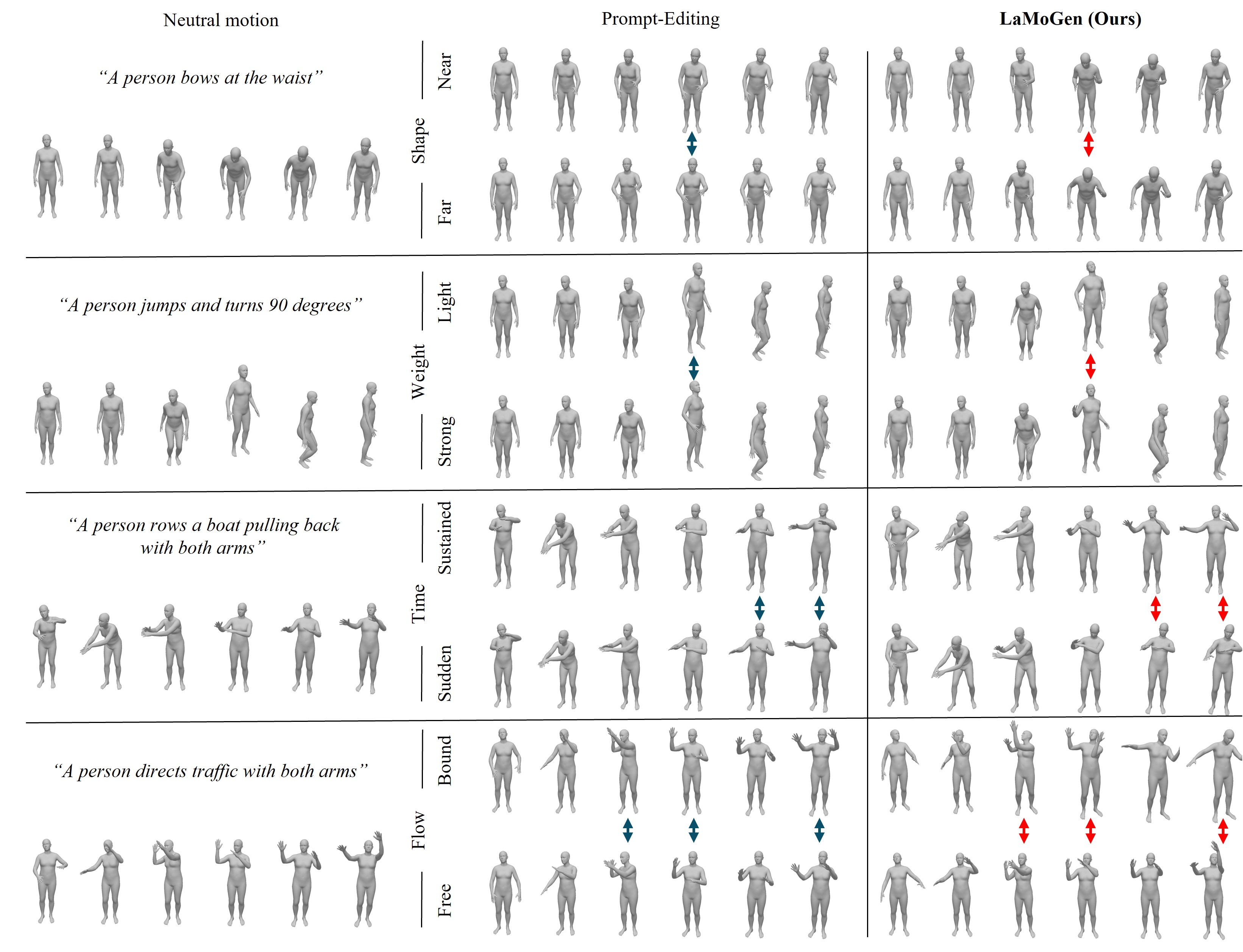}
	\vspace{-0.8em}
	\caption{Qualitative comparison of our method against the prompt editing baseline. For each attribute (Shape, Weight, Time, Flow), we show the motion generated from a neutral prompt (left), the result of prompt editing (middle), and the result from our proposed method (right). Our method produces distinct and controllable expressive variations that align with the target Laban components (red arrow), whereas the baseline shows minimal or incorrect changes (blue arrow). For Shape, the "Far" prompt results in a motion where the arms are spread wider and the head is bowed more deeply. Other descriptions are listed in the main text.}
	\label{fig:qualitative_comparison}
\end{figure*}

\section{Experiments}
We provide further details on our implementation, experiments, and additional qualitative results, including generated motion videos in the appendix.

\subsection{Implementation Details}
Our method is built upon MotionDiffuse~\cite{zhang2022motiondiffuse}, a transformer-based denoising diffusion model. For our guidance optimization, we employ the Adam optimizer with a learning rate of 5e-3. The Gaussian smoothing filter applied prior to each difference step uses a kernel size of 11 and $\sigma^2=10$. All experiments were conducted on a single NVIDIA A6000 GPU.

\subsection{Evaluation}
We conduct a series of experiments to evaluate the effectiveness of our proposed Laban-guided motion synthesis. We assess both qualitative and quantitative aspects of the generated motions, comparing our method against several strong baselines.

\paragraph{Baselines}
To the best of our knowledge, our work is the first to incorporate Laban Effort and Shape components for controllable text-to-motion synthesis. We compare our method against three baselines:
\begin{itemize}
    \item \textbf{Prompt Editing:} This baseline represents the most direct method for attribute control, appending the original text prompt with descriptive phrases. Following the descriptions in~\cite{samadani2020affective}, we use eight specific phrases corresponding to the extremes of the four Laban components to represent high and low intensities. The complete list of phrases is provided in the appendix.

    \item \textbf{Raw Frame Update:} This method applies an optimization loop directly to the final generated motion $x_0$. While model-agnostic, this approach optimizes the motion in the raw data space, which can easily lead to out-of-distribution and unnatural results as it does not leverage the learned manifold of the diffusion model.

    \item \textbf{Classifier Guidance:} A common technique for controllable generation. At each diffusion step $t$, the denoised sample $\hat{x}_0^t$ is updated using the gradient from a pre-trained Laban attribute classifier: $\hat{x}_{0}^{t'} = \hat{x}_{0}^t - \lambda \nabla \mathcal{L}_{\text{Laban}}(\hat{x}_{0}^t)$. The updated sample is then used to condition the next diffusion step.
\end{itemize}

\subsubsection{Qualitative Results}
Figure~\ref{fig:qualitative_comparison} presents a qualitative comparison, demonstrating that our Laban-guided approach generates motions that clearly reflect the intended expressive attributes. We highlight specific examples from the figure:
\begin{itemize}
    \item For \textbf{Shape}, the "Far" prompt results in a motion where the arms are spread wider and the head is bowed more deeply.
    \item For \textbf{Weight}, the "Strong" condition produces visibly more pronounced arm movements and conveys a greater sense of effort, particularly in the preparation and landing phases of a jump.
    \item For \textbf{Time}, "Sudden" prompts our model to generate a motion where the arms rise and fall faster, resembling a quick rowing action with more repeated strokes.
    \item For \textbf{Flow}, "Bound" creates a restrained movement with arm motions restricted to the front, whereas "Free" leads to expansive and liberated motions, such as reaching the arms high.
\end{itemize}

\begin{table*}[!t]
	\caption{Quantitative evaluation of controllability on the HumanML3D~\cite{guo2022generating} test set. We measure the mean relative change $\frac{f_f-f_i}{f_i}$ for each component when a specific attribute is targeted. Diagonality measures the disentanglement of control. This metric given a 4$\times$4 matrix A is calculated by $ \frac{\sum_i A_{ii}^2}{\sum_{i,j} A_{ij}^2} $.  Our method achieves the highest diagonality, indicating superior selective control.}
	\label{tab:controllability}
	\centering
	\begin{adjustbox}{width=1.0\linewidth}
		\begin{tabular}{llccccc}
			\toprule
			& & Weight & Time & Flow & Shape & Diagonality $\uparrow$ \\
			\midrule
			\multirow{4}{*}{Prompt Editing}
			& Weight (Light $\rightarrow$ Strong)   &  0.791$\pm$1.360 & 0.394$\pm$0.751 & 0.272$\pm$0.445 & 0.246$\pm$0.460 & \multirow{4}{*}{0.445} \\
			& Time (Sustained $\rightarrow$ Sudden)     & 0.235$\pm$1.038 & 0.065$\pm$0.480 & 0.005$\pm$0.285 & 0.017$\pm$0.272 &          \\
			& Flow (Bound $\rightarrow$ Free)    & 0.055$\pm$0.920 & -0.101$\pm$0.364 & -0.117$\pm$0.269 & -0.005$\pm$0.288 &         \\
			& Shape (Near $\rightarrow$ Far)    & 0.620$\pm$1.221 & 0.265$\pm$0.562 & 0.175$\pm$0.320 & 0.180$\pm$0.414 &\\
			
			\midrule
			\multirow{4}{*}{Raw Frame Update}
			& Weight (Light $\rightarrow$ Strong)   &  3.084$\pm$1.014 & 0.539$\pm$0.318 & -0.003$\pm$0.022 & 0.209$\pm$0.228 & \multirow{4}{*}{0.973} \\
			& Time (Sustained $\rightarrow$ Sudden)     & -0.085$\pm$0.237 & 0.179$\pm$0.244 & 0.018$\pm$0.027 & -0.040$\pm$0.111 &          \\
			& Flow (Bound $\rightarrow$ Free)    & 0.011$\pm$0.215 & 0.044$\pm$0.197 & 0.476$\pm$0.054 & 0.014$\pm$0.093 &         \\
			& Shape (Near $\rightarrow$ Far)    & -0.023$\pm$0.115 & -0.018$\pm$0.082 & -0.000$\pm$0.019 & 1.606$\pm$0.261 &\\
			
			\midrule
			\multirow{4}{*}{Classifier Guidance}
			& Weight (Light $\rightarrow$ Strong)   &  0.510$\pm$0.451 & 0.232$\pm$0.137 & 0.150$\pm$0.058 & 0.072$\pm$0.058 & \multirow{4}{*}{0.802} \\
			& Time (Sustained $\rightarrow$ Sudden)     & 0.048$\pm$0.039 & 0.047$\pm$0.037 & 0.044$\pm$0.022 & 0.011$\pm$0.009 &          \\
			& Flow (Bound $\rightarrow$ Free)    & 0.040$\pm$0.027 & 0.039$\pm$0.024 & 0.043$\pm$0.017 & 0.009$\pm$0.007 &         \\
			& Shape (Near $\rightarrow$ Far)    & 0.030$\pm$0.049 & 0.016$\pm$0.026 & 0.017$\pm$0.016 & 0.322$\pm$0.114 &\\
			
			\midrule
			\multirow{4}{*}{\textbf{Laban Guidance(ours)}}
			& Weight (Light $\rightarrow$ Strong)   &  3.081$\pm$2.033 & 0.323$\pm$0.369 & 0.023$\pm$0.046 & 0.163$\pm$0.305 & \multirow{4}{*}{\textbf{0.978}} \\
			& Time (Sustained $\rightarrow$ Sudden)     & 0.357$\pm$0.730 & 0.418$\pm$0.451 & 0.081$\pm$0.101 & 0.098$\pm$0.259 &          \\
			& Flow (Bound $\rightarrow$ Free)    & 0.065$\pm$0.716 & 0.029$\pm$0.371 & 0.379$\pm$0.116 & -0.032$\pm$0.214 &         \\
			& Shape (Near $\rightarrow$ Far)    & -0.044$\pm$0.307 & -0.040$\pm$0.153 & -0.012$\pm$0.032 & 1.613$\pm$0.470 &\\
			\bottomrule
		\end{tabular}
	\end{adjustbox}
\end{table*}

In stark contrast, the prompt editing baseline often fails to induce these nuanced changes, showing minimal to no significant differences from the neutral motion. In some instances, it even produces incorrect actions; for example, under the "Shape" condition, the figure fails to bend at the waist as intended, highlighting the unreliability of simple text manipulation for controlling expressive details.

\begin{figure}[t]
	\centering
	\includegraphics[width=1.0\linewidth]{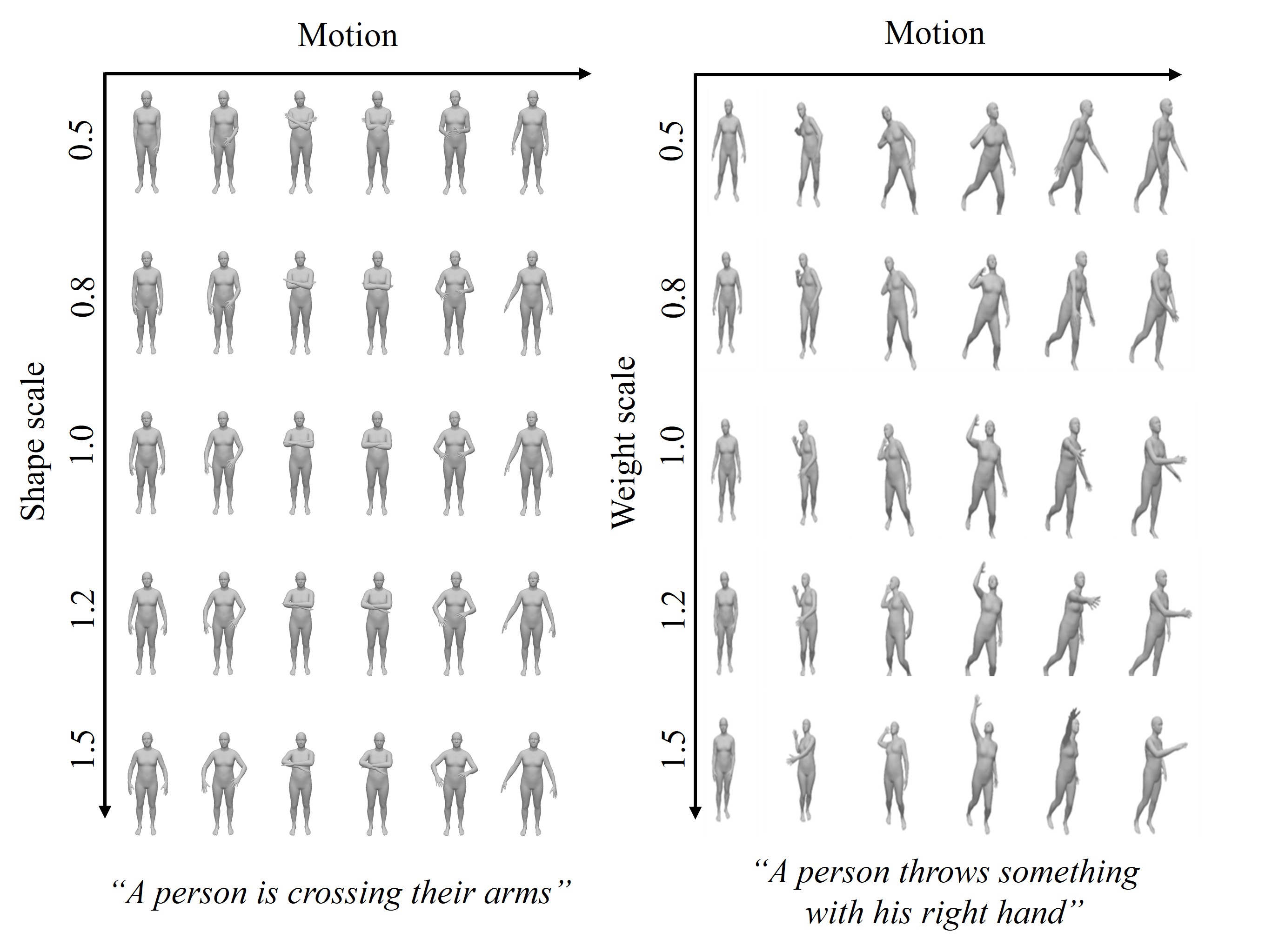}
	\vspace{-0.8em}
	\caption{Interpolation of motion expressiveness by varying the guidance scale for Shape and Weight attributes. Our method allows for smooth and continuous control over the intensity of the desired Laban component.}
	\label{fig:interpolation}
\end{figure}

\paragraph{Control Scale Interpolation}
To demonstrate the continuous controllability of our method, we interpolate the guidance scale factor with values $\{0.5, 0.8, 1.0, 1.2, 1.5\}$. As shown in Figure~\ref{fig:interpolation}, this smoothly transitions the motion expressiveness for both Shape and Weight attributes, confirming that our method provides fine-grained and reliable control over the generation process. 

\subsubsection{Quantitative Analysis}
\paragraph{Controllability and Disentanglement}
To quantitatively measure the controllability of our method, we evaluate how selectively each Laban component can be modified without affecting the others. For this analysis, we target a single Laban Effort component for modification and then compute the mean relative change induced across all four components. The comprehensive results of the mean and standard deviation are summarized in a 4$\times$4 change matrix A in Table~\ref{tab:controllability}, where diagonal elements represent the intended changes and off-diagonal elements represent unintended cross-effects. To distill this into a single score, we report the diagonality of this matrix, a metric for disentanglement. This metric, calculated as $\frac{\sum_i A_{ii}^2}{\sum_{i,j} A_{ij}^2}$,
intuitively represents the ratio of energy from desired, on-target changes to the total energy of all changes, including off-target side effects. A higher value signifies better disentanglement, indicating that a larger proportion of the modification is concentrated on the intended attribute. Our method achieves the highest diagonality score, demonstrating its superior ability to modify a specific expressive attribute while minimally disturbing the others.

\paragraph{Text-Motion Alignment and Fidelity}
We further evaluate the impact of our guidance on standard text-to-motion generation metrics: R-Precision, Frechet Inception Distance (FID), and Diversity. Table~\ref{tab:main_quantitative} shows that our method incurs a slight, expected trade-off in R-Precision and FID compared to the unguided baseline. This is an anticipated outcome, as steering the generation towards specific expressive attributes can naturally cause minor deviations from the original text-conditioned distribution. Nevertheless, our method maintains strong text correspondence, significantly outperforming the Raw Frame Update baseline and remaining competitive with other approaches while offering unparalleled expressive control.

\begin{table}[!htb]
	\caption{Comparison on standard text-to-motion metrics on the HumanML3D~\cite{guo2022generating} test set. Our method maintains competitive performance while providing fine-grained expressive control.}
	\label{tab:main_quantitative}
	\centering
	\begin{adjustbox}{width=1.0\linewidth}
		\begin{tabular}{lccccc}
			\toprule
			& \multicolumn{3}{c}{R-Precision $\uparrow$} & FID $\downarrow$ & Diversity $\uparrow$ \\
			\cmidrule(lr){2-4}
			Method & Top-1 & Top-2 & Top-3 & & \\
			\midrule
			MotionDiffuse~\cite{zhang2022motiondiffuse} & 0.470 & 0.662 & 0.767 & 2.21 & 9.32 \\
			\midrule
			Prompt Editing & 0.445 & 0.630 & 0.737 & 1.67 & 8.72 \\
			Raw Frame Update & 0.383 & 0.565 & 0.689 & 1.73 & 8.54 \\
			Classifier Guidance & 0.432 & 0.629 & 0.720 & 1.70 & 8.59 \\
			\textbf{Laban Guidance (Ours)} & 0.424 & 0.613 & 0.729 & 2.80 & 7.92 \\
			\bottomrule
		\end{tabular}
	\end{adjustbox}
\end{table}

\begin{figure}[!b]
	\centering
	\includegraphics[width=1.0\linewidth]{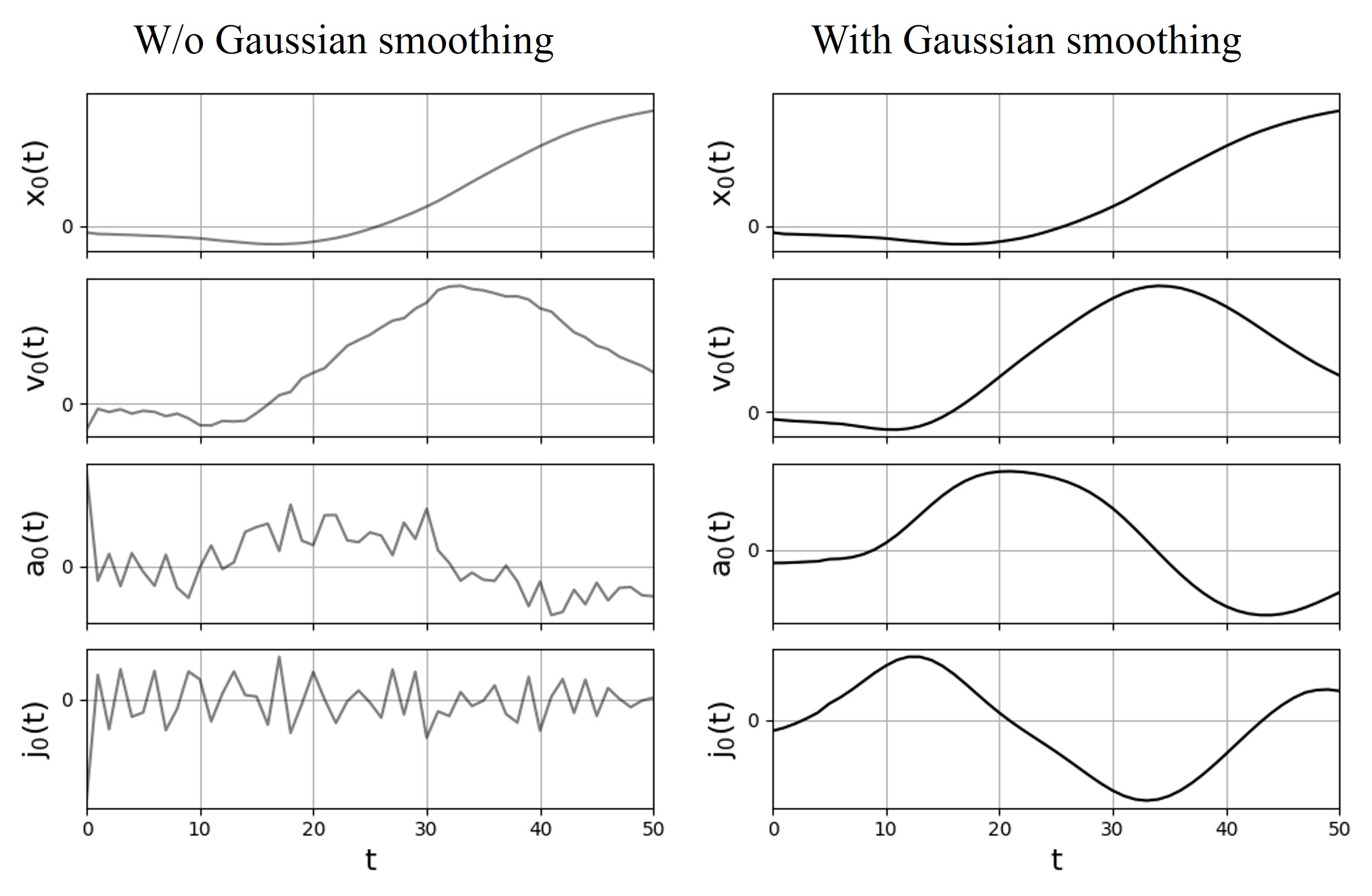}
	\vspace{-0.8em}
	\caption{Kinematics of the root joint over time, with and without Gaussian smoothing applied before differentiation. The plots for acceleration and jerk (third, fourth row) highlight the necessity of the smoothing process.}
	\label{fig6_smoothing}
\end{figure}

\subsection{Ablation Studies}
\paragraph{Effectiveness of Gaussian Smoothing}
We apply Gaussian smoothing to the motion data prior to computing time derivatives. This preprocessing step is crucial, as the derivative operation inherently amplifies high-frequency noise, which is inherent in the MotionDiffuse model. As illustrated in Figure~\ref{fig6_smoothing}, without smoothing, the resulting acceleration and jerk signals are dominated by noise artifacts. This renders the derived Time and Flow features, which rely on these derivatives, unreliable for guiding the generation process.

\begin{figure}[!b]
	\centering
	\includegraphics[width=0.85\linewidth]{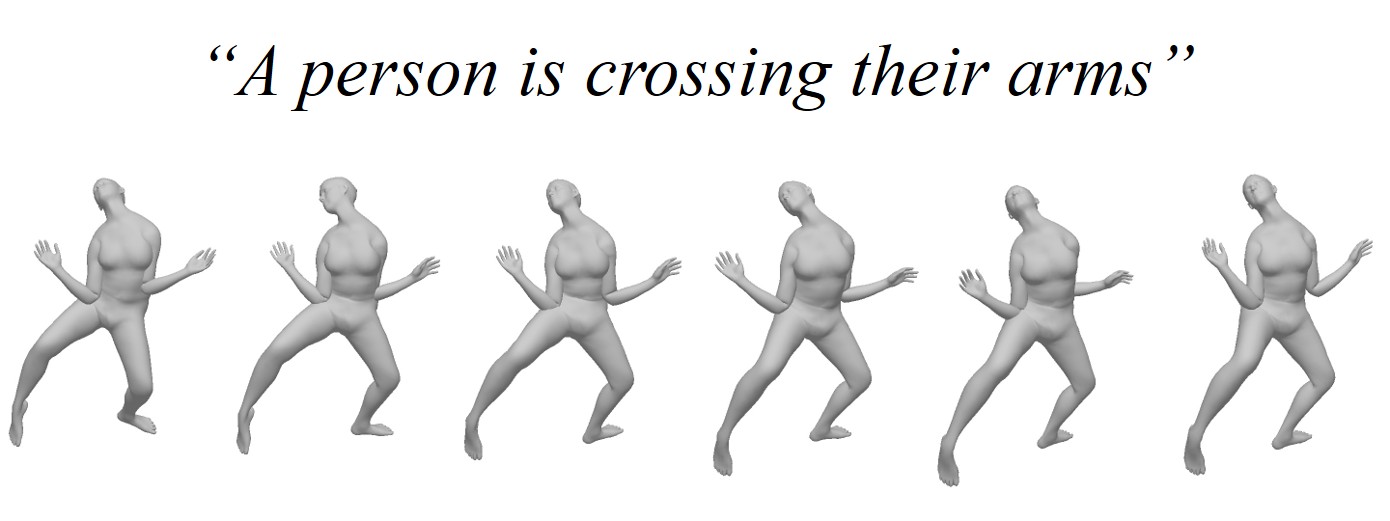}
	\vspace{-0.8em}
	\caption{An example of generation failure producing an implausible motion when the learning rate is excessively large.}
	\label{fig5_largelr}
\end{figure}

\paragraph{Effect of Learning Rate}
The learning rate for the embedding update is a critical hyperparameter. A sufficiently small learning rate results in negligible deviation from the original text-conditioned motion, failing to incorporate the Laban guidance. Conversely, an excessively large learning rate can lead to unstable updates, causing motion artifacts and generation failure, as depicted in Figure~\ref{fig5_largelr}. This analysis underscores the importance of selecting an appropriate learning rate to balance expressive control with motion plausibility.

\section{Discussion}
\paragraph{Limitations and Future Work}
Our work serves as a proof-of-concept for LMA-based motion control. A key avenue for future work is to move beyond our heuristic Laban mappings and learn a direct, perceptually-grounded mapping through user studies. This would also allow for a rigorous perceptual evaluation of the feature disentanglement suggested by our quantitative metrics. Finally, while our inference-time guidance is flexible, exploring how to distill this control into the model architecture for a more efficient single-pass generation remains an exciting research direction.

\section{Conclusion}
We introduce LaMoGen, a zero-shot method for Laban-guided text-to-motion generation. Our approach operates by iteratively refining the conditioning embedding during the DDIM sampling process, guided by a proposed Laban loss function at inference time. This allows LaMoGen to steer the generation process towards expressive motion qualities that are difficult to specify using text prompts alone, without requiring any additional training data. Our quantitative and qualitative evaluations demonstrate that the proposed method provides a high degree of expressive motion control, with a minor and acceptable trade-off in text-motion alignment.

\section{Acknowledgements}
This work was supported in part by Institute of Information \& communications Technology Planning \& Evaluation (IITP) grant funded by the Korea government(MSIT) [NO.RS-2021-II211343, Artificial Intelligence Graduate School Program (Seoul National University)] and by 2025 Student-Directed Education Regular Program from Seoul National University. Also, the authors acknowledged the financial support from the BK21 FOUR program of the Education and Research Program for Future ICT Pioneers, Seoul National University.

\bibliography{aaai2026}

\begin{thebibliography}{27}
\providecommand{\natexlab}[1]{#1}

\bibitem[{Ahuja and Morency(2019)}]{ahuja2019language2pose}
Ahuja, C.; and Morency, L.-P. 2019.
\newblock Language2pose: Natural language grounded pose forecasting.
\newblock In \emph{2019 International conference on 3D vision (3DV)}, 719--728.
  IEEE.

\bibitem[{Azadi et~al.(2023)Azadi, Shah, Hayes, Parikh, and
  Gupta}]{azadi2023make}
Azadi, S.; Shah, A.; Hayes, T.; Parikh, D.; and Gupta, S. 2023.
\newblock Make-An-Animation: Large-Scale Text-conditional 3D Human Motion
  Generation.
\newblock \emph{arXiv preprint arXiv:2305.09662}.

\bibitem[{Deng et~al.(2009)Deng, Dong, Socher, Li, Li, and Fei-Fei}]{imagenet}
Deng, J.; Dong, W.; Socher, R.; Li, L.-J.; Li, K.; and Fei-Fei, L. 2009.
\newblock ImageNet: A large-scale hierarchical image database.
\newblock In \emph{2009 IEEE Conference on Computer Vision and Pattern
  Recognition}, 248--255.

\bibitem[{Gleicher(2001)}]{gleicher2001motion}
Gleicher, M. 2001.
\newblock Motion path editing.
\newblock \emph{ACM Transactions on Graphics (TOG)}, 20(2): 114--139.

\bibitem[{Guo et~al.(2022)Guo, Zou, Zuo, Wang, Ji, Li, and
  Cheng}]{guo2022generating}
Guo, C.; Zou, S.; Zuo, X.; Wang, S.; Ji, W.; Li, X.; and Cheng, L. 2022.
\newblock Generating diverse and natural 3d human motions from text.
\newblock In \emph{Proceedings of the IEEE/CVF Conference on Computer Vision
  and Pattern Recognition}, 5152--5161.

\bibitem[{Guo et~al.(2020)Guo, Zuo, Wang, Zou, Sun, Deng, Gong, and
  Cheng}]{guo2020action2motion}
Guo, C.; Zuo, X.; Wang, S.; Zou, S.; Sun, Q.; Deng, A.; Gong, M.; and Cheng, L.
  2020.
\newblock Action2motion: Conditioned generation of 3d human motions.
\newblock In \emph{Proceedings of the 28th ACM International Conference on
  Multimedia}, 2021--2029.

\bibitem[{Ho, Jain, and Abbeel(2020)}]{ho2020denoising}
Ho, J.; Jain, A.; and Abbeel, P. 2020.
\newblock Denoising diffusion probabilistic models.
\newblock \emph{Advances in neural information processing systems}, 33:
  6840--6851.

\bibitem[{Ho and Salimans(2022)}]{ho2022classifier}
Ho, J.; and Salimans, T. 2022.
\newblock Classifier-free diffusion guidance.
\newblock \emph{arXiv preprint arXiv:2207.12598}.

\bibitem[{Holden, Saito, and Komura(2016)}]{holden2016deep}
Holden, D.; Saito, J.; and Komura, T. 2016.
\newblock A deep learning framework for character motion synthesis and editing.
\newblock In \emph{ACM TOG}, volume~35, 1--11. ACM.

\bibitem[{Kim et~al.(2009)Kim, Kim, You, and Oh}]{kim2009stable}
Kim, S.; Kim, C.; You, B.; and Oh, S. 2009.
\newblock Stable whole-body motion generation for humanoid robots to imitate
  human motions.
\newblock In \emph{2009 IEEE/RSJ International Conference on Intelligent Robots
  and Systems}, 2518--2524. IEEE.

\bibitem[{Laban and Ullmann(1971)}]{laban1971mastery}
Laban, R.; and Ullmann, L. 1971.
\newblock The mastery of movement.

\bibitem[{LaViers et~al.(2016)LaViers, Bai, Bashiri, Heddy, and
  Sheng}]{LaViers2016}
LaViers, A.; Bai, L.; Bashiri, M.; Heddy, G.; and Sheng, Y. 2016.
\newblock \emph{Abstractions for Design-by-Humans of Heterogeneous Behaviors},
  237--262.
\newblock Cham: Springer International Publishing.
\newblock ISBN 978-3-319-25739-6.

\bibitem[{Lin et~al.(2023)Lin, Zeng, Lu, Cai, Zhang, Wang, and
  Zhang}]{lin2023motionx}
Lin, J.; Zeng, A.; Lu, S.; Cai, Y.; Zhang, R.; Wang, H.; and Zhang, L. 2023.
\newblock Motion-X: A Large-scale 3D Expressive Whole-body Human Motion
  Dataset.
\newblock \emph{Advances in Neural Information Processing Systems}.

\bibitem[{Mousas(2018)}]{mousas2018performance}
Mousas, C. 2018.
\newblock Performance-driven dance motion control of a virtual partner
  character.
\newblock In \emph{2018 IEEE Conference on Virtual Reality and 3D User
  Interfaces (VR)}, 57--64. IEEE.

\bibitem[{Petrovich, Black, and Varol(2022)}]{petrovich2022temos}
Petrovich, M.; Black, M.~J.; and Varol, G. 2022.
\newblock Temos: Generating diverse human motions from textual descriptions.
\newblock In \emph{European Conference on Computer Vision}, 480--497. Springer.

\bibitem[{Samadani, Gorbet, and Kulic(2020)}]{samadani2020affective}
Samadani, A.; Gorbet, R.; and Kulic, D. 2020.
\newblock Affective movement generation using Laban effort and shape and hidden
  markov models.
\newblock \emph{arXiv preprint arXiv:2006.06071}.

\bibitem[{Samadani et~al.(2013)Samadani, Burton, Gorbet, and
  Kulic}]{samadani2013laban}
Samadani, A.-A.; Burton, S.; Gorbet, R.; and Kulic, D. 2013.
\newblock Laban effort and shape analysis of affective hand and arm movements.
\newblock In \emph{2013 Humaine Association conference on affective computing
  and intelligent interaction}, 343--348. IEEE.

\bibitem[{Song, Meng, and Ermon(2020)}]{song2020denoising}
Song, J.; Meng, C.; and Ermon, S. 2020.
\newblock Denoising diffusion implicit models.
\newblock \emph{arXiv preprint arXiv:2010.02502}.

\bibitem[{Tevet et~al.(2022)Tevet, Raab, Gordon, Shafir, Cohen-Or, and
  Bermano}]{tevet2022human}
Tevet, G.; Raab, S.; Gordon, B.; Shafir, Y.; Cohen-Or, D.; and Bermano, A.~H.
  2022.
\newblock Human motion diffusion model.
\newblock \emph{arXiv preprint arXiv:2209.14916}.

\bibitem[{Wan et~al.(2023)Wan, Dou, Komura, Wang, Jayaraman, and
  Liu}]{wan2023tlcontrol}
Wan, W.; Dou, Z.; Komura, T.; Wang, W.; Jayaraman, D.; and Liu, L. 2023.
\newblock TLControl: Trajectory and Language Control for Human Motion
  Synthesis.
\newblock \emph{arXiv preprint arXiv:2311.17135}.

\bibitem[{Xie et~al.(2023)Xie, Li, Huang, Liu, Zhang, Zheng, and
  Shou}]{xie2023boxdiff}
Xie, J.; Li, Y.; Huang, Y.; Liu, H.; Zhang, W.; Zheng, Y.; and Shou, M.~Z.
  2023.
\newblock BoxDiff: Text-to-Image Synthesis with Training-Free Box-Constrained
  Diffusion.
\newblock \emph{arXiv preprint arXiv:2307.10816}.

\bibitem[{Xie et~al.(2024)Xie, Jampani, Zhong, Sun, and
  Jiang}]{xie2024omnicontrol}
Xie, Y.; Jampani, V.; Zhong, L.; Sun, D.; and Jiang, H. 2024.
\newblock OmniControl: Control Any Joint at Any Time for Human Motion
  Generation.
\newblock In \emph{The Twelfth International Conference on Learning
  Representations}.

\bibitem[{Yuan et~al.(2022)Yuan, Song, Iqbal, Vahdat, and
  Kautz}]{yuan2022physdiff}
Yuan, Y.; Song, J.; Iqbal, U.; Vahdat, A.; and Kautz, J. 2022.
\newblock Physdiff: Physics-guided human motion diffusion model.
\newblock \emph{arXiv preprint arXiv:2212.02500}.

\bibitem[{Zhang, Rao, and Agrawala(2023)}]{zhang2023adding}
Zhang, L.; Rao, A.; and Agrawala, M. 2023.
\newblock Adding conditional control to text-to-image diffusion models.
\newblock In \emph{Proceedings of the IEEE/CVF international conference on
  computer vision}, 3836--3847.

\bibitem[{Zhang et~al.(2022)Zhang, Cai, Pan, Hong, Guo, Yang, and
  Liu}]{zhang2022motiondiffuse}
Zhang, M.; Cai, Z.; Pan, L.; Hong, F.; Guo, X.; Yang, L.; and Liu, Z. 2022.
\newblock Motiondiffuse: Text-driven human motion generation with diffusion
  model.
\newblock \emph{arXiv preprint arXiv:2208.15001}.

\bibitem[{Zhang et~al.(2023)Zhang, Guo, Pan, Cai, Hong, Li, Yang, and
  Liu}]{zhang2023remodiffuse}
Zhang, M.; Guo, X.; Pan, L.; Cai, Z.; Hong, F.; Li, H.; Yang, L.; and Liu, Z.
  2023.
\newblock Remodiffuse: Retrieval-augmented motion diffusion model.
\newblock In \emph{Proceedings of the IEEE/CVF International Conference on
  Computer Vision}, 364--373.

\bibitem[{Zhong et~al.(2024)Zhong, Xie, Jampani, Sun, and
  Jiang}]{zhong2024smoodi}
Zhong, L.; Xie, Y.; Jampani, V.; Sun, D.; and Jiang, H. 2024.
\newblock Smoodi: Stylized motion diffusion model.
\newblock In \emph{European Conference on Computer Vision}, 405--421. Springer.

\end{thebibliography}

\newpage
\appendix
\begin{center}
{\Large \bf
Appendix \\
\vspace{0.5em}
}
\end{center}

\subsection{Implementation Details}
\subsubsection{Pretrained Model Details}
Our framework is built upon MotionDiffuse~\cite{zhang2022motiondiffuse} model which is publicly available in https://github.com/mingyuan-zhang/MotionDiffuse. The model architecture is a transformer-based network trained on the HumanML3D dataset~\cite{guo2022generating}. The key architectural parameters are summarized in Table~\ref{tab:model_params}. The motion data is represented as a sequence of 263-dimensional vectors, including joint rotations, positions, and velocities, at 20 FPS. The text embedding $\mathbf{e}$ subject to update is input to a stylization Block, which is interleaved between attention layers.

\begin{table}[h]
\centering
\caption{Architectural details of the pretrained MotionDiffuse model.}
\label{tab:model_params}
\begin{tabular}{lc}
\toprule
Parameter & Value \\
\midrule
Number of Transformer Layers & 8 \\
Number of Attention Heads & 8 \\
Latent Dimension & 512 \\
Text Embedding Dimension & 2048 \\
Motion Data Dimension & 263 \\
\bottomrule
\end{tabular}
\end{table}

\subsubsection{Detailed LMA Quantification Formulas}
As mentioned in the main paper, we adapt the LMA quantification from \citet{samadani2020affective}. The motion kinematics (velocity $\vect{v}$, acceleration $\vect{a}$, jerk $\vect{j}$) are computed from the joint positions $\vect{x}$ using first-order finite differences:
\begin{align}
    \vect{v}_t &= \vect{x}_t - \vect{x}_{t-1} \\
    \vect{a}_t &= \vect{v}_t - \vect{v}_{t-1} \\
    \vect{j}_t &= \vect{a}_t - \vect{a}_{t-1}
\end{align}
Before computing these differences, we apply a Gaussian smoothing filter (kernel size=11, $\sigma^2=10$) to the position data $\vect{x}$ to mitigate noise amplification.

The instantaneous LMA features at frame $t$ for a set of end-effectors $\mathcal{E}$ are then defined as:
\begin{itemize}
    \item \textbf{Weight ($E_t$):} $\sum_{k \in \mathcal{E}} m_k \|\vect{v}_{k,t}\|^2$ (where $m_k$ is the mass of end-effector $k$, assumed to be 1 for simplicity).
    \item \textbf{Time ($A_t$):} $\sum_{k \in \mathcal{E}} \|\vect{a}_{k,t}\|$
    \item \textbf{Flow ($J_t$):} $\sum_{k \in \mathcal{E}} \|\vect{j}_{k,t}\|$
    \item \textbf{Shape ($V_t$):} The volume of the 3D axis-aligned bounding box of the entire skeleton at frame $t$.
\end{itemize}

\subsubsection{Hyperparameters}
For the embedding optimization during guidance, we perform $K=1$ optimization step per DDIM sampling step. The Adam optimizer uses a learning rate of $0.005$ and betas $(\beta_1, \beta_2) = (0.7, 0.9)$. The stability constant $\delta$ in the Laban loss is set to $\delta=1 \times 10^{-6}$. We used 1000 DDIM steps for sampling.\\
For the raw frame update baseline, we applied $K=100$ optimization step to the output motion using Adam optimizer with learning rate of $0.05$ and betas $(\beta_1, \beta_2) = (0.7, 0.9)$.\\
For the classifier guidance baseline, we applied $K=1$ optimization step with learning rate of $0.005$.

\subsection{Experimental Details}
\subsubsection{Laban Tag to Scale Mappings}
The scalar values for each Laban tag was determined heuristically to produce a perceptible effect. The default values used for each tag are listed in Table~\ref{tab:scale_maps}.

\begin{table}[h]
\centering
\caption{Laban Tag to Scale Mappings}
\label{tab:scale_maps}
\begin{adjustbox}{width=0.7\linewidth}
\begin{tabular}{lll}
\toprule
LMA Component & Small Value & Large Value \\
\midrule
Weight & 0.5 & 1.5 \\
Time & 0.8 & 1.2 \\
Flow & 0.8 & 1.2 \\
Shape & 0.5 & 1.5 \\
\bottomrule
\end{tabular}
\end{adjustbox}
\end{table}

\subsubsection{Full List of Phrases for Prompt Editing Baseline}
Table~\ref{tab:prompt_phrases} lists the full set of descriptive phrases used for the Prompt Editing baseline. These phrases were appended to the original text prompts to represent the high and low intensities of the four LMA components.

\begin{table}[h]
\centering
\caption{Descriptive phrases used for the Prompt Editing baseline.}
\label{tab:prompt_phrases}
\begin{adjustbox}{width=\linewidth}
\begin{tabular}{lll}
\toprule
LMA Component & Small Value & Large Value \\
\midrule
Weight & "with small energy" & "with large energy" \\
Time & "in a sustained time" & "in a quick time" \\
Flow & "bounded" & "freely" \\
Shape & "with small volume" & "with large volume" \\
\bottomrule
\end{tabular}
\end{adjustbox}
\end{table}

\subsubsection{Evaluation Metrics Details}
To ensure a fair comparison, we follow the standard evaluation protocol for text-to-motion generation. R-Precision, FID, and Diversity scores were computed using the official evaluation code and pretrained feature extractors provided by \citet{guo2022generating} for the HumanML3D dataset. The feature extractor is a motion-text joint embedding model that maps both modalities into a shared latent space.

\begin{figure}[b]
	\centering
	\includegraphics[width=1.0\linewidth]{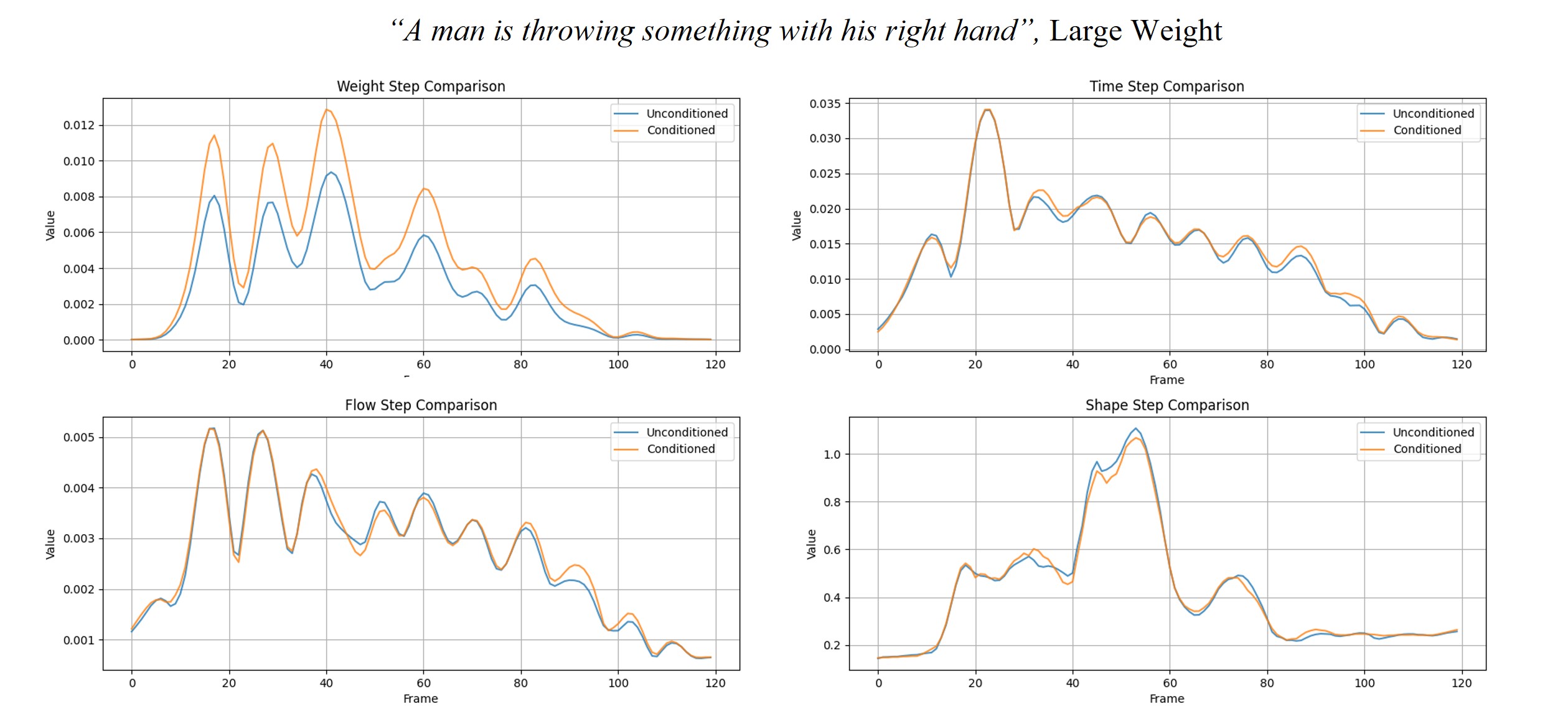}
	\vspace{-0.8em}
	\caption{Comparison of Text only Conditioned Generation (blue line) and Fully Conditioned Generation (orange line)}
	\vspace{-1.5em}
	\label{fig:appendix_tc_fc}
\end{figure}

\subsection{Comparison of Text only Conditioned Generation and Fully Conditioned Generation}
Figure~\ref{fig:appendix_tc_fc} shows the calculated Laban features for the motion generated in Step 1(text conditioned generation) and Step 2(fully conditioned generation). The graphs each represent a Laban effort component. The Weight feature increased by a factor for each frame, and other features remained the same, demonstrating controllable generation.

\subsection{Additional Qualitative Results}
\begin{figure*}[h]
    \centering
    \includegraphics[width=0.9\linewidth]{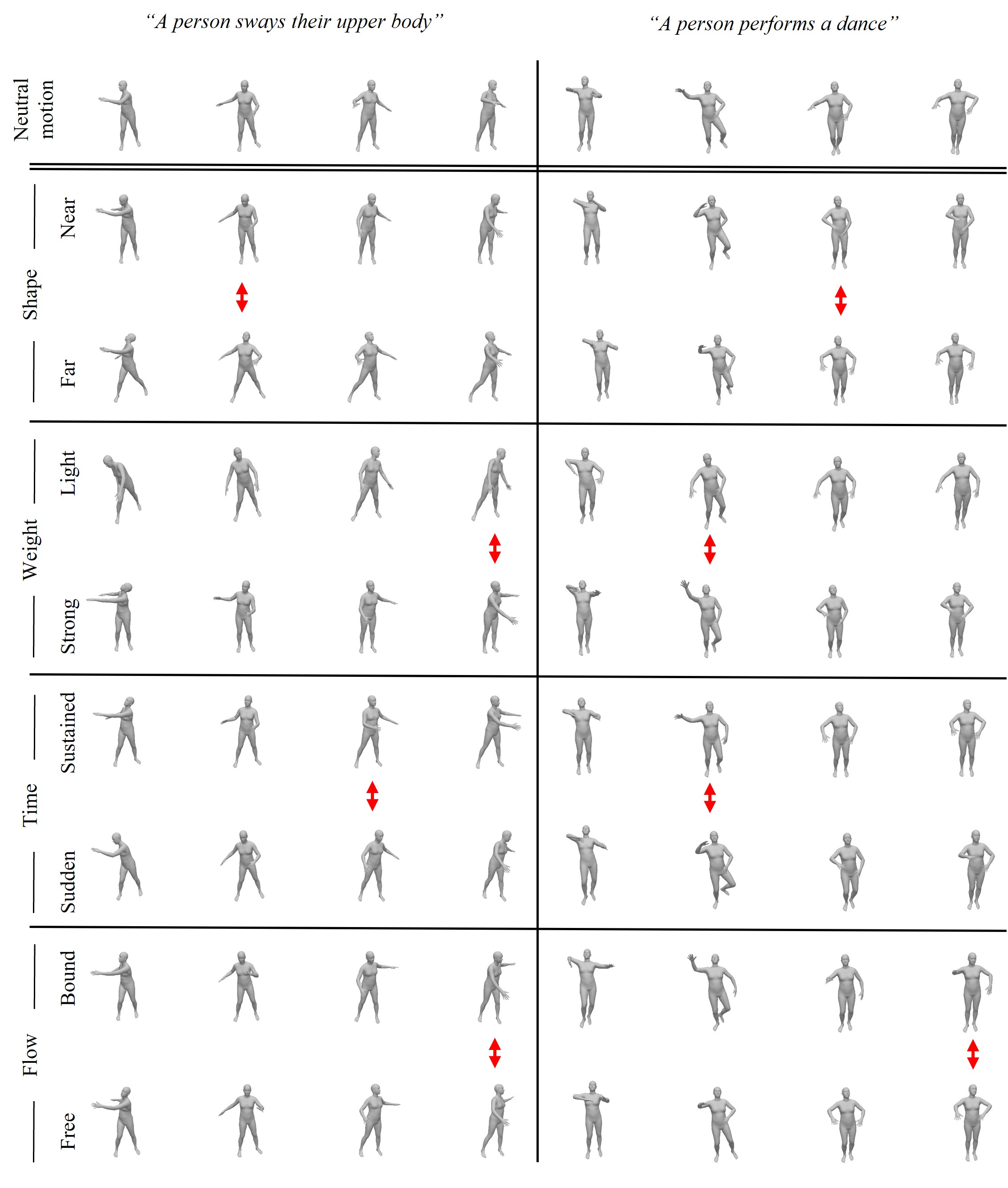}
    \caption{Additional qualitative results.  Our method successfully modulates the expressive LMA targets. Frames with notable difference are marked with red arrows.}
    \label{fig:appendix_more_examples1}
\end{figure*}

\begin{figure*}[h]
    \centering
    \includegraphics[width=0.9\linewidth]{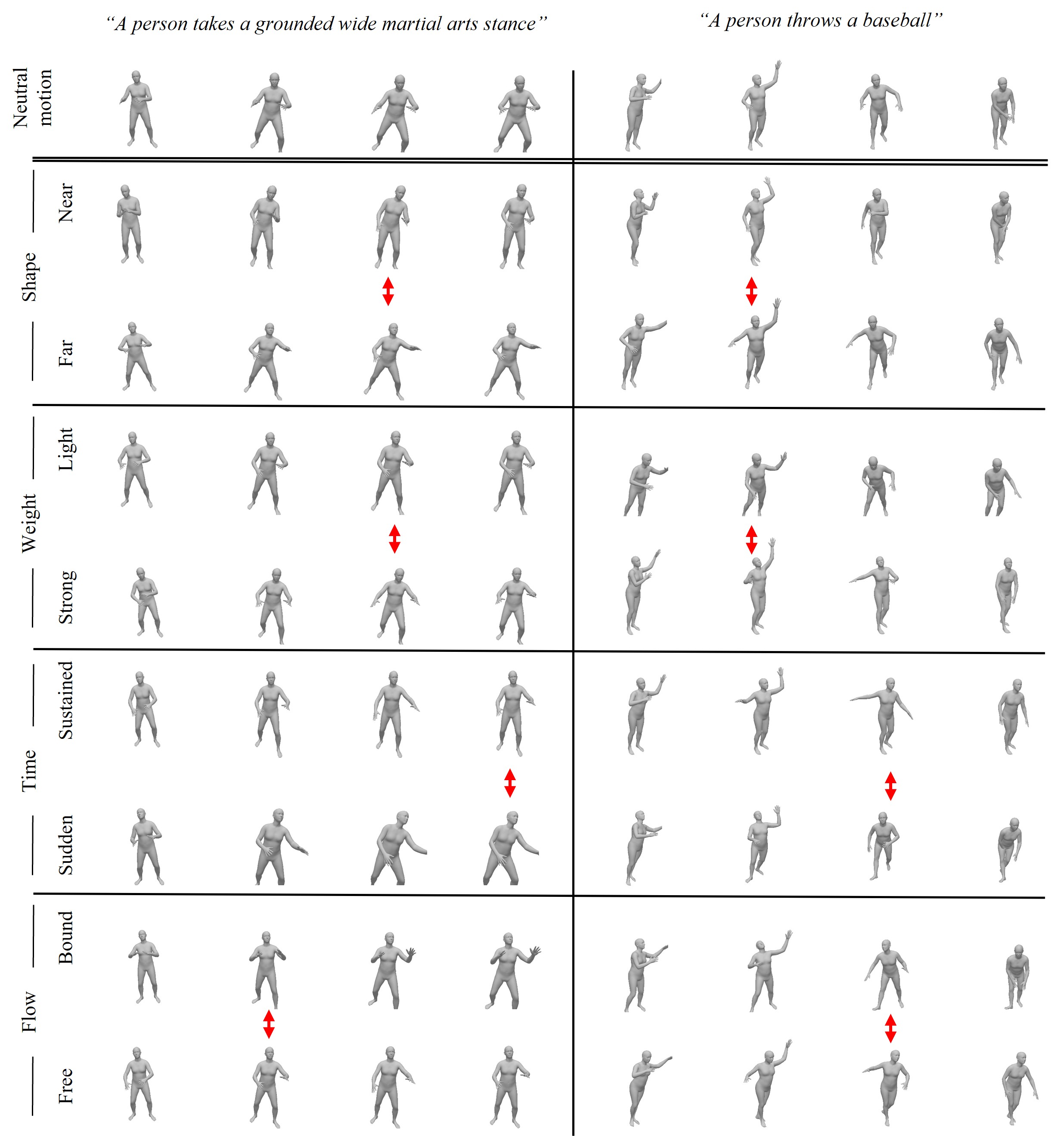}
    \caption{Additional qualitative results.  Our method successfully modulates the expressive LMA targets. Frames with notable difference are marked with red arrows.}
    \label{fig:appendix_more_examples2}
\end{figure*}

Figure~\ref{fig:appendix_more_examples1} and Figure~\ref{fig:appendix_more_examples2} at next page shows additional qualitative results for different text prompts, demonstrating the generalizability of our method. For each prompt, we show the effect of controlling the Shape, Weight, Time, and Flow components using our method.

\end{document}